\RequirePackage[x11names]{xcolor}
\documentclass[conference,a4paper]{IEEEtran}
\IEEEoverridecommandlockouts

\usepackage[hidelinks]{hyperref}
\usepackage[cmex10]{amsmath}
\usepackage{amssymb,amsfonts}
\usepackage{dblfloatfix}

\usepackage[ruled,vlined]{algorithm2e}
\usepackage{graphicx}
\graphicspath{{Figures/PDF/}{Figures/PNG/}}
\usepackage{multirow}

\usepackage{booktabs}
\usepackage{siunitx}
\usepackage[numbers,compress]{natbib}
\usepackage{texnames}
\usepackage{bm,bbm}
\usepackage{orcidlink}

\usepackage{pifont}
\usepackage{booktabs}
\usepackage{colortbl}

\usepackage{todonotes}

\newcommand{\cmark}{\ding{51}} 
\newcommand{\xmark}{\ding{55}} 

\begin{document}



\title{AGI for the Earth, the path, possibilities and how to evaluate intelligence of models that work with Earth Observation Data? }

\author{\IEEEauthorblockN{Mojtaba Valipour\orcidlink{0000-0002-5877-2869}, Kelly Zheng, James Lowman\orcidlink{0000-0002-1745-9454}, Spencer Szabados, Mike Gartner, and Bobby Braswell\orcidlink{0000-0002-4061-9516}}
	\IEEEauthorblockA{HUM.AI\\
		\{mojtaba, kelly\}@hum.ai}
}

\maketitle
\begin{abstract}
	Artificial General Intelligence (AGI) is closer than ever to becoming a reality, sparking widespread enthusiasm in the research community to collect and work with various modalities, including text, image, video, and audio. Despite recent efforts, satellite spectral imagery, as an additional modality, has yet to receive the attention it deserves. This area presents unique challenges, but also holds great promise in advancing the capabilities of AGI in understanding the natural world. In this paper, we argue why Earth Observation data is useful for an intelligent model, and then we review existing benchmarks and highlight their limitations in evaluating the generalization ability of foundation models in this domain. This paper emphasizes the need for a more comprehensive benchmark to evaluate earth observation models. To facilitate this, we propose a comprehensive set of tasks that a benchmark should encompass to effectively assess a model's ability to understand and interact with Earth observation data. \footnote{Copyright 2024 IEEE. Published in the 2025 IEEE International Geoscience and Remote Sensing Symposium (IGARSS 2025), scheduled for 3 - 8 August 2025 in Brisbane, Australia. Personal use of this material is permitted. However, permission to reprint/republish this material for advertising or promotional purposes or for creating new collective works for resale or redistribution to servers or lists, or to reuse any copyrighted component of this work in other works, must be obtained from the IEEE. Contact: Manager, Copyrights and Permissions / IEEE Service Center / 445 Hoes Lane / P.O. Box 1331 / Piscataway, NJ 08855-1331, USA. Telephone: + Intl. 908-562-3966.}
\end{abstract}






\section{Introduction}


Artificial General Intelligence (AGI) represents the next frontier in the evolution of artificial intelligence, aiming to replicate human-like cognitive abilities across diverse domains \cite{bubeck2023sparks, morris2023levels}. Unlike narrow AI, which excels in specific tasks but lacks adaptability, AGI seeks to possess the flexibility and understanding to perform any intellectual task that a human can. As the world faces increasingly complex challenges, from climate change and environmental degradation\footnote{\href{https://knowledge4policy.ec.europa.eu/climate-change-environmental-degradation_en}{knowledge4policy.ec.europa.eu}} to natural disasters and resource management, the potential of AGI to process vast amounts of data \cite{hoffmann2022training} and make informed decisions is increasingly critical. However, achieving the full potential of AGI requires not only advances in algorithmic techniques but also the availability of diverse \cite{miranda2023beyond}, and high-quality data \cite{hoffmann2022training}.



 Most AGI research efforts to date have primarily focused on modalities like text \cite{achiam2023gpt, dubey2024llama}, largely driven by vast, publicly available datasets such as Common Crawl\footnote{commoncrawl.org}. Human language only encapsulates one dimension of intelligence in which we can encode observations, events, their interrelations, and basically any concept that can be described textually. The innovation of current large language models (LLMs) is in their excellent ability to reason about a wide range of concepts and tasks outside the scope of the data used to train them; that is, these models are notable for their great generalization performance over a wide variety of tasks. More recently, other modalities such as audio \cite{yang2023uniaudio}, image \cite{ho2020denoising}, and videos \cite{ho2022video, agarwal2025cosmos}, where the diversity of the dataset is important \cite{miranda2023beyond}

There are many reasons why Earth observation (EO) may be beneficial or even necessary for AGI. Put simply, language is limited to things that have been written down \cite{floridi2020gpt3}. There is a relationship between written knowledge and what is represented in the vast volumes of EO data, and where they overlap, they should be mostly consistent. For example, people have written about what happens to soils when you grow a monoculture. However, no one has written down how every square meter of the Earth processes material and energy in its own idiosyncratic way. One can say, "Tropical soils under these conditions do X", but an understanding of the behavior of the Earth system must have geospatial and temporal specificity in order to be predictive. Moreover, some emerging research questions whether we will run out of human-generated text data for training intelligent models \cite{villalobosposition}. We believe Earth-Observation is potentially an important part of creating levels 3 to 5 of the general intelligence \cite{morris2023levels}.

We posit that an analogous framework can be developed based on learning the underlying processes of the biological and physical Earth system, using the vast and diverse corpus of observational data available from remote observations and other sources (EO data). This would enable a wide range of decision intelligence applications, similar to how LLMs allow users to seek a solution to a problem by providing contextual information sufficient to trigger practical and usable responses. In the case of an Earth intelligence model, rather than human language, the means of interaction necessarily will be sequences of encoded observations or observation-like values. The joint distributions of all these possible quantitative values, whether they are reflectance, backscatter, temperature, etc., are essentially the ``vocabulary'' of the model. For interpretability, more control and easier communications between these agents and humans, one can align the representations of these low-level features with a large language model \cite{liu2024remoteclip}.

Foundation models in EO and remote sensing are rapidly evolving \cite{xiao2024foundationmodelsremotesensing}, \cite{rs17020179}, with pre-trained models being fine-tuned for various tasks such as land cover classification \cite{helber2019eurosat, Karra2021CoverMapping, Brown2022CoverMapping}, object detection \cite{Li_2020}, and environmental monitoring. Convolutional Neural Networks (CNNs), such as U-Net \cite{ronneberger2015u} and ResNet \cite{he2016deep}, traditionally have shown great success in tasks like image segmentation and land cover mapping. The more recent Transformer-based models like Vision Transformers (ViT) \cite{dosovitskiy2020image}, HSViT \cite{xu2024hsvit}, Segment-Anything Model (SAM) \cite{kirillov2023segment, osco2023segment, zhou2024mesam}, Maseked Auto-Encoders (MAE) \cite{he2022masked}, Swin Transformer \cite{liu2021swin}, SpectralGPT \cite{hong2024spectralgpt}, Prithvi \cite{szwarcman2024prithvi}, DOFA \cite{xiong2024neural}, One For All \cite{xiong2024one}, DETR \cite{carion2020end} and diffusion-based models like DiffusionSat \cite{khanna2023diffusionsat}, and TerraMind \cite{jakubik2025terramind} are also gaining traction in satellite imagery. Transformers are known for capturing long-range spatial dependencies and learning rich representations. In comparison, diffusion-based models are more suitable for image-generative tasks thanks to their concurrent generation. These foundation models are advancing towards generalized, temporal, satellite-agnostic, and multi-task learning frameworks, holding great promise for AGI applications in Earth observation.




 As these models become more available, the importance of having high-quality benchmarks for evaluating them becomes more evident. We currently have many well-known high-quality benchmarks for assessing the performance of AGI on non-earth-observation data, including, but not limited to \cite{chollet2024arc, hendrycks2020measuring, jimenez2023swe, chan2024mle, laurent2024lab, chao2024jailbreakbench}. However, despite the recent efforts, such a list for evaluating the performance of AGI on earth-observation data is nothing but short. This is because most of the available benchmarks are targeting either traditional non-generative EO tasks or they are super targeted, as we analyzed and showed in this paper \cite{xiong2022earthnets, marsocci2024pangaea, paolo2022xview3sardetectingdarkfishing, yeh2021sustainbench}.

In the following sections, we introduce a new categorical system to analyze the available benchmarks and set of tasks that we think should be considered for the future of earth observation evaluation. Then, we discuss current benchmarks and their limitations. Finally, we advocate for new application-oriented benchmarks designed to capture Earth Observation models' generalization ability.



\begin{table*}[ht]
\centering
\caption{Earth Observation Task Coverage by Benchmark}
\resizebox{\textwidth}{!}{%
\begin{tabular}{llccccccccccc}
\toprule
\textbf{Category} & \textbf{Task} &
\rotatebox{90}{\textbf{PANGAEA [42]}} &
\rotatebox{90}{\textbf{EarthNets [41]}} &
\rotatebox{90}{\textbf{xView3 [43]}} &
\rotatebox{90}{\textbf{SustainBench [44]}} &
\rotatebox{90}{\textbf{GEO-Bench [63]}} &
\rotatebox{90}{\textbf{PhilEO [73]}} &
\rotatebox{90}{\textbf{FoMo [74]}} &
\rotatebox{90}{\textbf{VRSBench [87]}} &
\rotatebox{90}{\textbf{UrBench [91]}} &
\rotatebox{90}{\textbf{VLEO-Bench [92]}} &
\textbf{All} \\
\midrule
\multirow{10}{*}{Sat2Info} & \cellcolor{LightYellow1}Pattern/Anomaly Detection & \xmark & \xmark & \xmark & \xmark & \xmark & \xmark & \xmark & \xmark & \xmark & \xmark & \xmark \\
& \cellcolor{AntiqueWhite1}Trend Forecasting & \xmark & \xmark & \xmark & \xmark & \xmark & \xmark & \xmark & \xmark & \xmark & \xmark & \xmark \\
& \cellcolor{Azure3}Change Detection / Classification / Quantification & \cellcolor{Azure3}\cmark & \xmark & \xmark & \cellcolor{Azure3}\cmark & \xmark & \xmark & \xmark & \xmark & \xmark & \cellcolor{Azure3}\cmark & \cellcolor{Azure3}\cmark \\
& \cellcolor{Aquamarine1}Visual Question Answering & \xmark & \xmark & \xmark & \xmark & \xmark & \xmark & \xmark & \cellcolor{Aquamarine1}\cmark & \cellcolor{Aquamarine1}\cmark & \cellcolor{Aquamarine1}\cmark & \cellcolor{Aquamarine1}\cmark \\
& \cellcolor{DarkSlateGray3}Scene Understanding / Captioning & \xmark & \xmark & \xmark & \xmark & \xmark & \xmark & \xmark & \cellcolor{DarkSlateGray3}\cmark & \xmark & \cellcolor{DarkSlateGray3}\cmark & \cellcolor{DarkSlateGray3}\cmark \\
& \cellcolor{DarkSeaGreen2}Scene Regression / Estimation & \cellcolor{DarkSeaGreen2}\cmark & \xmark & \cellcolor{DarkSeaGreen2}\cmark & \cellcolor{DarkSeaGreen2}\cmark & \xmark & \cellcolor{DarkSeaGreen2}\cmark & \xmark & \xmark & \xmark & \xmark & \cellcolor{DarkSeaGreen2}\cmark \\
& \cellcolor{OliveDrab3}Object Detection / Tracking & \xmark & \cellcolor{OliveDrab3}\cmark & \cellcolor{OliveDrab3}\cmark & \xmark & \xmark & \xmark & \cellcolor{OliveDrab3}\cmark & \cellcolor{OliveDrab3}\cmark & \cellcolor{OliveDrab3}\cmark & \cellcolor{OliveDrab3}\cmark & \cellcolor{OliveDrab3}\cmark \\
& \cellcolor{SpringGreen3}Segmentation / Categorization & \cellcolor{SpringGreen3}\cmark & \cellcolor{SpringGreen3}\cmark & \xmark & \cellcolor{SpringGreen3}\cmark & \cellcolor{SpringGreen3}\cmark & \cellcolor{SpringGreen3}\cmark & \cellcolor{SpringGreen3}\cmark & \xmark & \xmark & \cellcolor{SpringGreen3}\cmark & \cellcolor{SpringGreen3}\cmark \\
& \cellcolor{SeaGreen4}Geo-Localization & \xmark & \xmark & \xmark & \xmark & \xmark & \xmark & \xmark & \xmark & \cellcolor{SeaGreen4}\cmark & \cellcolor{SeaGreen4}\cmark & \cellcolor{SeaGreen4}\cmark \\
& \cellcolor{Cyan4}Scene Classification & \xmark & \cellcolor{Cyan4}\cmark & \cellcolor{Cyan4}\cmark & \cellcolor{Cyan4}\cmark & \cellcolor{Cyan4}\cmark & \cellcolor{Cyan4}\cmark & \cellcolor{Cyan4}\cmark & \xmark & \xmark & \cellcolor{Cyan4}\cmark & \cellcolor{Cyan4}\cmark \\
\midrule
\multirow{6}{*}{Data2Sat} & \cellcolor{DeepSkyBlue1}Scene Retrieval and Search & \xmark & \xmark & \xmark & \xmark & \xmark & \xmark & \xmark & \xmark & \cellcolor{DeepSkyBlue1}\cmark & \xmark & \cellcolor{DeepSkyBlue1}\cmark \\
& \cellcolor{RoyalBlue1}Scenario Generation & \xmark & \xmark & \xmark & \xmark & \xmark & \xmark & \xmark & \xmark & \xmark & \xmark & \xmark \\
& \cellcolor{Gold1}Scene Generation & \xmark & \xmark & \xmark & \xmark & \xmark & \xmark & \xmark & \xmark & \xmark & \xmark & \xmark \\
& \cellcolor{Tan1}Sensor Replication & \xmark & \xmark & \xmark & \xmark & \xmark & \xmark & \xmark & \xmark & \xmark & \xmark & \xmark \\
& \cellcolor{Chocolate1}Counterfactual Analysis & \xmark & \xmark & \xmark & \xmark & \xmark & \xmark & \xmark & \xmark & \xmark & \xmark & \xmark \\
& \cellcolor{DeepPink1}Intervention & \xmark & \xmark & \xmark & \xmark & \xmark & \xmark & \xmark & \xmark & \xmark & \xmark & \xmark \\
\midrule
\multirow{1}{*}{Sat2Model} & \cellcolor{OliveDrab1}2D/3D Scene Reconstruction & \xmark & \xmark & \xmark & \xmark & \xmark & \xmark & \xmark & \xmark & \xmark & \xmark & \xmark \\
\midrule
\multirow{6}{*}{Sat2Sat} & \cellcolor{MistyRose1}Scene Interpolation & \xmark & \xmark & \xmark & \xmark & \xmark & \xmark & \xmark & \xmark & \xmark & \xmark & \xmark \\
& \cellcolor{Pink1}Scene Editing / Object Removal & \xmark & \xmark & \xmark & \xmark & \xmark & \xmark & \xmark & \xmark & \xmark & \xmark & \xmark \\
& \cellcolor{Plum2}Style Transfer / Satellite Mapping & \xmark & \xmark & \xmark & \xmark & \xmark & \xmark & \xmark & \xmark & \xmark & \xmark & \xmark \\
& \cellcolor{Orchid1}Super Resolution & \xmark & \xmark & \xmark & \xmark & \xmark & \xmark & \xmark & \xmark & \xmark & \cellcolor{Orchid1}\cmark & \cellcolor{Orchid1}\cmark \\
& \cellcolor{Magenta1}Data Fusion & \xmark & \xmark & \xmark & \xmark & \xmark & \xmark & \xmark & \xmark & \xmark & \xmark & \xmark \\
& \cellcolor{MediumPurple1}Scene Prediction / Possible Futures & \xmark & \xmark & \xmark & \xmark & \xmark & \xmark & \xmark & \xmark & \xmark & \cellcolor{MediumPurple1}\cmark & \cellcolor{MediumPurple1}\cmark \\
\bottomrule
\end{tabular}%
}
\label{tab:eo_grouped_rotated}
\end{table*}

\section{Methodology}


We first present a framework for categorizing the primary satellite-related data tasks of interest, addressing their diversity and current benchmark limitations. The framework defines four key categories based on input-output requirements and their role in measuring intelligence. 


There are many ways to categorize Earth Observation tasks. Our categorization is focused on diversity and on informing approaches to model generalization. It is intended to support the research and development of algorithms, models, and systems by serving as a structured framework from which robust, diverse benchmarks can be constructed.

\textbf{Satellite to Information (Sat2Info)}: tasks that involve extracting broad types of information from satellite images such as geographic attributes, metadata, and higher-level data derived from raw imagery like land usage patterns or environmental indicators. 

    \begin{itemize}
        \item \colorbox{LightYellow1}{Pattern/Anomaly Detection}: Pattern detection identifies recurring features in satellite images. Anomaly detection spots unusual deviations from regular patterns. For example, detecting storms or identifying oil spills. 
        \item \colorbox{AntiqueWhite1}{Trend Forecasting}: Predict future changes based on historical satellite data, such as forecasting urban growth or deforestation by analyzing patterns over time. 
        \item \colorbox{Azure3}{Change Detection/Classification/Quantification}: Change detection identifies differences in images over time, Classification categorizes the changes (e.g., deforestation). Quantification measures their extent (e.g., hectares of forest lost). 
        \item \colorbox{Aquamarine1}{Visual Question Answering}: Ask questions about satellite images, such as “How many parks are visible?” and generate answers based on image content. 
        \item \colorbox{DarkSlateGray3}{Scene Understanding/Captioning}: Generate descriptive text for satellite images, such as “An urban area with dense residential buildings,” summarizing key features. 
        \item \colorbox{DarkSeaGreen2}{Scene Regression/Estimation}: Predict continuous variables like temperature, biomass or soil moisture from satellite imagery, aiding environmental monitoring and forecasting. 
        \item \colorbox{OliveDrab3}{Object Detection/Tracking \& Visual Grounding}: Locate and track specific objects based on natural language expression, queries or labels. Object/Instance Detection can locate a building, and Tracking follows its movement over time. Visual Grounding associates a query like “find the river” with the corresponding object in the image. 
        \item \colorbox{SpringGreen3}{Segmentation/Categorization}: Segmentation divide an image into regions, and categorization assigns labels, such as distinguishing between urban, forest, and agricultural land. 
        \item \colorbox{SeaGreen4}{Geo-Localization}: Determine the exact location (e.g. GPS coordinates) of objects in satellite images, enabling accurate mapping and navigation, like pinpointing new buildings in a city. 
        \item \colorbox{Cyan4}{Scene Classification}: Assign labels to satellite images based on content, such as classifying an image as urban, forest, or agricultural land. 
    \end{itemize}
    

    
\textbf{Data to Satellite (Data2Sat)}: tasks that involve generating satellite-like imagery from non-satellite data or synthetic distributions. This includes using GIS or climate data to create synthetic satellite images (e.g., simulating urban growth) or transforming random noise into realistic satellite imagery using generative models like GANs\cite{goodfellow2014generative}, Transformers\cite{vaswani2017attention}, and Diffusion Models \cite{ho2020denoising} (e.g., generating images similar to those from Sentinel-2 using Landsat data). 

    \begin{itemize}
        \item \colorbox{DeepSkyBlue1}{Scene Retrieval and Search}: Find similar satellite images based on a query, and scene search allows locating specific images based on visual features or descriptions. 
        \item \colorbox{RoyalBlue1}{Scenario Generation}: Create synthetic satellite images to simulate possible environmental conditions, such as forest areas under different climate change scenarios. 
        \item \colorbox{Gold1}{Scene Generation}: Create new satellite images from data inputs, like simulating a new urban development using existing city imagery. 
        \item \colorbox{Tan1}{Sensor Replication}: Mimic the output of a specific satellite sensor, such as generating images similar to those from Sentinel-2 using a data source other than satellite data, for example a specific geo-location. 
        \item \colorbox{Chocolate1}{Counterfactual Analysis}: Generate alternate versions of satellite images by altering specific factors, like simulating urban growth under different policies. 
        \item \colorbox{DeepPink1}{Intervention}: Modify satellite data or model parameters to test the impact of changes, such as simulating deforestation or reforestation. 
    \end{itemize}
    
\textbf{Satellite to Model (Sat2Model)}: Sat2Model focuses on the use of satellite imagery as input for the generation of predictive models or simulations. These models could be applied in a variety of fields, such as climate modeling, agricultural forecasting, or urban growth prediction. Tasks under this category involve extracting relevant features from satellite images and using them to build models that can forecast future conditions or outcomes.

    \begin{itemize}
        \item \colorbox{OliveDrab1}{2D/3D Scene Reconstruction}: 2D/3D reconstruction creates images/models from satellite images, like building a 3D city model for urban planning or environmental impact assessments. 
    \end{itemize}
    
\textbf{Satellite to Satellite (Sat2Sat)}: tasks that involve the transformation of one satellite image into another, often across different time points, sensors, or modalities. This includes tasks like image registration, translation, fusion, and temporal change detection. Sat2Sat is particularly useful in monitoring changes over time, such as tracking deforestation, urban sprawl, or the effects of natural disasters.

    \begin{itemize}
        \item \colorbox{MistyRose1}{Scene Interpolation}: Generate intermediate images between two time/location points, useful for visualizing changes when data from intermediate times are unavailable. 
        \item \colorbox{Pink1}{Scene Editing/Object Removal}: Modify satellite images, such as removing cloud cover to reveal underlying land features for a more clear analysis. 
        \item \colorbox{Plum2}{Style Transfer/Satellite Mapping}: Change the appearance of satellite images, like simulating daylight on nighttime images. Satellite mapping aligns images from different sensors to standardize data. 
        \item \colorbox{Orchid1}{Super Resolution}: Enhance the resolution of satellite images, revealing finer details, such as improving image quality for better analysis of urban areas. 
        \item \colorbox{Magenta1}{Data Fusion}: Combine different satellite data sources (e.g., optical, radar) to create more accurate and detailed representations of an area. 
        \item \colorbox{MediumPurple1}{Scene Prediction/Possible Futures}: Forecast future satellite images by considering various possible scenarios based on current data. Scene Prediction typically focuses on predicting the most likely future scenario (e.g., urban sprawl or environmental changes). Possible Futures explores multiple potential outcomes based on different assumptions or interventions (e.g., the effects of various climate policies). 
    \end{itemize}


The categorization is designed to reflect the significant tasks that are common in satellite imagery applications or essential for a generative model, while avoiding unnecessary category overlap. Categories like Satellite to Features (Sat2Features) or Satellite to Distribution (Sat2Dist), which could be seen as a subset of Sat2Info, are intentionally excluded to maintain simplicity and clarity. Sat2Sat and Data2Sat were kept separate to emphasize the importance of Sat2Sat tasks. Further AGI-specific tasks relevant in the near future may include continuous learning, transfer learning, and generalization across domains. We acknowledge that this is not a comprehensive list.

\section{Benchmarks}


In this section, we analyzed the available benchmarks in the literature using our proposed methodology. 

As shown in Table \ref{tab:eo_grouped_rotated}, despite the growing interest in applying machine learning to Earth observation (EO), our analysis reveals that most existing benchmarks are heavily skewed toward narrow tasks, primarily within the Sat2Info category. Tasks such as scene classification, object detection, and segmentation are well represented across benchmarks like EarthNets [41], PANGAEA [42], and VLEO-Bench [92]. However, even within this dominant category, higher-order semantic tasks such as visual question answering, scene captioning, and regression remain inconsistently supported. Moreover, tasks related to spatial reasoning (e.g., geo-localization) and temporal pattern analysis (e.g., trend forecasting) are virtually absent from all current benchmarks.

More notably, entire categories such as Data2Sat (e.g., scene generation, counterfactual reasoning, intervention modeling) and Sat2Model (e.g., 2D/3D reconstruction) are effectively unrepresented. This gap indicates that current EO benchmarks are not yet designed to evaluate or encourage generalizable or generative intelligence — core capabilities of AGI systems. While a few datasets like UrBench [91] and VRSBench [87] introduce grounded visual tasks, they are limited in scope and lack integration with broader reasoning objectives. Thus, there remains a critical need for comprehensive, multi-modal benchmarks that evaluate compositional understanding, causality, and generation across space, time, and sensor modalities — capabilities that future AGI systems must master.

We propose the development of a standard and comprehensive yet simple and intuitive Earth Observation benchmark that covers a wide range of tasks essential for evaluating foundation models. We believe all the tasks in table \ref{tab:eo_grouped_rotated} should be included in a comprehensive benchmark. This benchmark would not only include fundamental tasks such as scene classification and segmentation but also extend to more complex tasks like scene generation, counterfactual analysis, and scenario prediction. By including both generative and analytical tasks, the benchmark would enable the evaluation of models’ ability to create new data, understand context, and make informed predictions or interventions. Additionally, tasks like data fusion and scene interpolation would assess a model’s capacity to handle multi-modal, temporal, and spatial data. The goal is to provide a balanced framework that is both accessible for the community and robust enough to drive advancements in practical Earth Observation applications. We envision this as a crucial step for future work, aimed at fostering the development of models capable of tackling challenges in sustainability, climate change, and disaster response.

\section{Conclusion}

In conclusion, the development of Artificial General Intelligence (AGI) for Earth Observation holds immense potential for tackling global challenges such as environmental degradation, climate change, and disaster response. This paper emphasizes the need for a comprehensive benchmark to evaluate AGI models, focusing on a wide range of tasks that go beyond basic understanding and incorporate the complexities of real-world Earth Observation scenarios. While existing benchmarks address specific tasks, they do not fully capture the generalizability, reasoning, and creativity required for broader AGI applications. We propose the creation of an intuitive and balanced benchmark that spans categories like scene classification, change detection, scene generation, counterfactual analysis, and more. However, several downstream tasks lack sufficient labeled data for meaningful evaluation. Thus, we encourage the community to collaborate in developing this benchmark, ensuring it is comprehensive and adaptable for various datasets and geographies. This collective effort will help drive the progress of AGI models, making them more capable and effective in addressing pressing environmental and sustainability challenges.

\small
\bibliographystyle{IEEEtranN}
\bibliography{references}

\end{document}